\documentclass[3p, preprint, 12pt]{elsarticle}

\usepackage{graphicx}
\usepackage{amssymb}

\usepackage{lineno}
\usepackage{threeparttable}
\usepackage{amsmath}
\usepackage{mathtools}
\usepackage{bm}
\usepackage{subcaption}
\usepackage{booktabs}
\usepackage{hyperref}
\usepackage{multirow}
\usepackage{xcolor}
\usepackage[ruled,vlined]{algorithm2e}
\usepackage{hyperref}
\usepackage{enumitem}
\usepackage{makecell}
\usepackage{xcolor}

\begin{document}

\begin{frontmatter}


%
\title{Deep Trip Generation with Graph Neural Networks for \\Bike Sharing System Expansion}

\author[a]{Yuebing Liang} 
\author[a]{Fangyi Ding}
\author[a]{Guan Huang}
\author[a,b,c]{Zhan Zhao\corref{cor1}}

\address[a]{Department of Urban Planning and Design, The University of Hong Kong, Hong Kong SAR, China}
\address[b]{Urban Systems Institute, The University of Hong Kong, Hong Kong SAR, China}
\address[c]{Musketeers Foundation Institute of Data Science, The University of Hong Kong, Hong Kong SAR, China}

\cortext[cor1]{Corresponding author (zhanzhao@hku.hk)}

\begin{abstract}
Bike sharing is emerging globally as an active, convenient, and sustainable mode of transportation. To plan successful bike-sharing systems (BSSs), many cities start from a small-scale pilot and gradually expand the system to cover more areas. For station-based BSSs, this means planning new stations based on existing ones over time, which requires prediction of the number of trips generated by these new stations across the whole system. Previous studies typically rely on relatively simple regression or machine learning models, which are limited in capturing complex spatial relationships. Despite the growing literature in deep learning methods for travel demand prediction, they are mostly developed for short-term prediction based on time series data, assuming no structural changes to the system. In this study, we focus on the trip generation problem for BSS expansion, and propose a graph neural network (GNN) approach to predicting the station-level demand based on multi-source urban built environment data. Specifically, it constructs multiple localized graphs centered on each target station and uses attention mechanisms to learn the correlation weights between stations. We further illustrate that the proposed approach can be regarded as a generalized spatial regression model, indicating the commonalities between spatial regression and GNNs. The model is evaluated based on realistic experiments using multi-year BSS data from New York City, and the results validate the superior performance of our approach compared to existing methods. We also demonstrate the interpretability of the model for uncovering the effects of built environment features and spatial interactions between stations, which can provide strategic guidance for BSS station location selection and capacity planning.
\end{abstract}


\end{frontmatter}


\section{Introduction} \label{sec:intro}

Bike sharing is an emerging mode of transportation that is growing rapidly in many metropolitan areas around the world. It has proven to benefit our cities and societies in a number of ways, including reducing traffic congestion, enhancing inter-modal connections, alleviating air pollution and promoting healthier lifestyle \citep{shaheen2010bikesharing}. Due to these positive effects of bike sharing, many cities have invested on the deployment of bike sharing systems (BSSs) for urban residents, usually starting from a smaller scale first and gradually expanding over the years based on the demand potential. For station-based BSSs, this means building new stations based on existing ones. For instance, when New York City (NYC) first launched its BSS (called Citi Bike) in 2013, there were 329 stations in the city center, while by the end of 2019 the station network has expanded to the outskirt regions, reaching 882 stations. When planning for station-based BSS expansion, the prediction of potential demand for newly added stations is paramount for city planners and bike sharing service providers to make strategic decisions. Particularly, the knowledge of the expected number of trips originating or destined for each station can be used to support important planning decisions regarding when and where a new station should be built, how much capacity the station should have and how many bikes need to be allocated \citep{liu2017functional}. This is essentially the goal of trip generation, as the first step in the widely used four-step travel demand forecasting process, but specifically for bike sharing \citep{noland2016bikeshare}. In this study, we focus on the problem of trip generation for station-based bike sharing system expansion (TG-BSSE).

Due to the worldwide proliferation of bike sharing, increasing attention has been paid to bike sharing demand prediction. Traditionally, researchers use statistical regression models to capture the relationship between bike demand and surrounding geographic and demographic characteristics \citep{buck2012bike, bachand2012better}. To consider spatial dependencies between stations, spatial regression has been applied, whose main idea is to encode the space as a representation vector into the framework of regression statistics \citep{faghih2016incorporating, zhang2017exploring}. Although these regression methods have the clear advantage of good interpretability, they may be unable to accurately capture the real structure of demand patterns due to oversimplified (e.g., linear) assumptions. While improved performance has been achieved by recent studies using tree-based ensemble learning models for TG-BSSE \citep{liu2017functional, kouincorporating, guidon2020expanding}, these methods may not be flexible or expressive enough to recover the complexity of spatial interactions between BSS stations.

With rapid advances in machine learning and AI, deep neural networks (DNNs) have been applied to BSS demand prediction in recent years because of their capability to extract relationships in any arbitrary function forms. However, existing deep learning methods are mostly developed for short-term prediction in a mature and stable BSS, which uses historical demand sequences as input to predict demand in the near future (i.e., at most 24 hours) \citep{xu2018station, chai2018bike, li2019learning, liang2022cross}. They generally assume no structural changes to the system, rely on stationary demand data, and thus are not applicable to TG-BSSE. Though limited, several recent research works have applied DNNs for trip generation in new transportation sites by aggregating historical demand patterns of nearby existing sites \citep{gong2020potential, zhou2021modeling}. Another group of studies focused on dynamic demand prediction for a continuously evolving system \citep{luo2019dynamic,he2020dynamic}, which is regarded as a spatiotemporal prediction problem with historical demand sequences of existing stations as model input. Despite these recent attempts, there still exist several important research gaps to be addressed:

\begin{itemize}[noitemsep]
    \item First, existing deep learning models typically rely on sequential dependencies in the temporal dimension. However, for BSS expansion, the addition of new stations would likely disrupt the demand patterns for the entire system. In addition, considering the long planning time horizon of system expansion, the demand patterns can also change due to various exogenous factors (e.g., the opening of a new subway station). Therefore, models dependent on past sequential/temporal dependencies may not be suitable for BSS expansion. 
    \item Second, while changes in BSS network are generally less costly and more frequent compared to other mass transportation systems (e.g., metro), they are still rare events, occurring once every few months or years and often in batches. As a result, only a very limited number of network configurations can be observed in the historical data, leading to data sparsity issues and making it difficult to effectively learn spatial dependencies for trip generation.
    \item Third, the experiment design of most existing studies cannot validate their model effectiveness in real-world expansion scenarios, as their experiments are based on simulation data with short time intervals (i.e., from 30 minutes to 1 day). In practice, the planning and implementation of BSS expansion can take months or even years. Furthermore, the effect of new stations on the whole system is highly complex and can hardly be approximated with relatively simple simulations. 
    \item Fourth, although previous research has demonstrated the superior prediction performance of DNNs for demand prediction tasks (mostly in the short term), there is a lack of discussion regarding why the model makes such prediction. In the case of TG-BSSE, interpretability is crucial to gain a deeper understanding of the determinants of BSS demand for a specific station, which can provide valuable policy implications for future BSS network design.  
\end{itemize}

To address these research gaps, this study tackles the TG-BSSE problem by exploiting spatial dependencies and utilizing multi-source data to capture the effect of the urban built environment, including POIs, road networks, public transit facilities and socio-demographic information. Specifically, we propose a spatially-dependent multi-graph attention network (Spatial-MGAT) approach to incorporate various built environment features as well as heterogeneous spatial dependencies between stations. For each BSS station, the proposed model can effectively leverage its spatial dependencies on other stations that are either geographically close by or share similar built environment characteristics. The relationship between our proposed graph neural network (GNN) approach and the classic spatial regression model is further discussed. Experiments are conducted on a real-world multi-year BSS expansion dataset from New York City, and the results validate the effectiveness of our approach. Its interpretability is also demonstrated using explainable AI techniques. The main contributions of this paper are summarized as follows:

\begin{itemize}[noitemsep]
    \item This study makes one of the first attempts at using deep learning techniques to enhance trip generation for station-based BSS expansion based on multi-source urban built environment data. The proposed model focuses on learning spatial dependencies instead of temporal/sequential dependencies, making it more generalizable for different network configurations. Therefore, it can be used as a planning tool to provide strategic guidance for BSS network design and update.
    \item To capture spatial interactions across the network, we construct two localized graphs centered on each station based on geographical proximity and urban environment similarity respectively, and use attention mechanisms to adaptively learn correlation weights between connected stations. The use of localized graphs makes it possible to use each station as the analysis unit, which mitigates the data sparsity issue.
    \item We demonstrate that our proposed approach can be regarded as a generalized spatial regression model with nonlinear activation functions, heterogeneous spatial dependencies and adaptive spatial weights learned from data. This allows us to conceptually link spatial regression with GNNs, and provide another example of enhancing classic econometric models with state-of-the-art deep learning techniques.
    \item To validate the effectiveness of our proposed approach, the Citi Bike system in NYC is used as a case study. Extensive experiments are conducted based on multi-year data to approximate real-world BSS expansion scenarios. The results validate the superior performance of the proposed model against existing methods for both newly planned and existing stations. Further analysis demonstrates the ability of our approach to explain the determinants of BSS demand and BSS station interactions. 
\end{itemize}

\section{Literature Review} \label{sec:literature}

In this section, we first review existing works related to trip generation for BSS expansion, and then present a short summary of deep learning approaches for BSS demand prediction. In addition, we review recent research that uses deep neural networks to enhance traditional theory-based models in transportation research, which will be relevant to our methodology. 

\subsection{Trip Generation for Bike Sharing System Expansion} \label{literature:new}

Early investigation regarding the impact of the built environment on bike sharing trip generation was mainly based on statistical regression models. Linear regression, one of the simplest forms of regression models, has been commonly used to discover the determinants of BSS usage \citep{bachand2012better, rixey2013station}. Later research employed multi-level mixed regression models to capture dependencies between repeated observations from the same station \citep{faghih2014land,el2017effects}. To incorporate spatial dependencies between stations, \cite{faghih2016incorporating} and \cite{zhang2017exploring} leveraged spatial regression models and achieved better model fit. Thanks to the good interpretability of regression models, these studies have identified a large group of factors that are strongly associated with BSS station demand, including land use (e.g., CBD, restaurants, commercial enterprises) \citep{faghih2014land}, transportation facilities (e.g.,rail stations, nearby bicycle lanes) \citep{noland2016bikeshare}, socio-demographic features (e.g., job and population density, race, income) \citep{bachand2012better} and BSS network design (e.g. proximity to other BSS stations, distance to the center of BSS) \citep{rixey2013station}. However, these methods perform relatively poorly for actual demand forecasting, largely because the assumption of linear relationships between BSS usage and input features limits their ability to capture potentially more complex patterns underlying the data.

With their rapid advancement in recent years, machine learning methods have also been considered as a solution to TG-BSSE. \cite{guidon2020expanding} employed a tree-based ensemble method, namely Random Forest, to predict the demand for expanding BSS to a new city. \cite{kouincorporating} used XGBoost to incorporate spatial network information for improved model performance. A hierarchical demand prediction model was developed by \cite{liu2017functional}, which first clusters stations with similar POI characteristics and close geographical distances into functional zones, and then predicts BSS demands from functional zone level to station level using Random Forest and Ridge Regression. A hybrid approach was also adopted by \cite{hyland2018hybrid}, combining clustering techniques with regression modeling. However, these models only rely on built environment features of the target BSS station itself, and are limited in leveraging the spatial dependencies between stations.

DNNs have proven to be a powerful tool for capturing complex nonlinear relationships hidden in human mobility data. However, most existing DNN methods focus on short-term demand prediction for stable systems, as we will discuss in Section~\ref{literature:exist}. Only recently did several studies introduce DNNs to human mobility prediction for transportation planning applications. \cite{gong2020potential} proposed a multi-view localized correlation learning method, whose core idea is to learn the passenger flow correlations between the target areas and their localized areas with adaptive weights. \cite{zhou2021modeling} estimated the potential crowd flow at a newly planned site by leveraging the historical demand patterns of nearby multi-modal sites in a collective way. Another group of relevant studies considered demand prediction of time-varying transportation networks. \cite{luo2019dynamic} proposed a graph sequence learning approach for a rapidly expanding electric vehicle system. \cite{he2020dynamic} predicted the demand flow of an evolving dockless e-scooter system using a spatiotemporal graph capsule neural network. Their problem formulation, which is essentially a time-series prediction problem with historical demand series of existing transportation sites as model input, is quite different from ours. In real-world system expansion scenarios with long planning time horizon and potentially major network changes, such problem formulation can easily suffer from demand distribution discrepancy and data sparsity problems. Existing works often sidestepped these challenges by considering short time intervals from 30 minutes to 1 day or masking stations from one network snapshot to simulate possible system expansion, which cannot reflect the planning and deployment of real-world transportation system expansion over time.

\subsection{Deep Learning Methods for Bike Demand Prediction} \label{literature:exist}

In recent years, extensive studies have demonstrated the power of deep learning models for short-term prediction of BSS demand. Early research used recurrent neural networks (RNNs) to capture temporal dependencies in historical demand series. A long-short term memory neural network (LSTM) was used in \cite{xu2018station} considering exogenous factors (e.g., weather). \cite{zhang2018short} leveraged historical demand of public transit to enhance the prediction performance of BSS demand with LSTM. To incorporate spatial information, \cite{zhou2018predicting} and \cite{qiao2021dynamic} combined RNNs with convolutional neural networks (CNNs), which however, can only provide demand prediction at the grid cell level and are not suitable for station-based BSSs. To capture spatial dependencies across graph-structured data, GNNs have recently been employed for BSS demand prediction at the station level. \cite{lin2018predicting} proposed a graph learning framework with GCNs and RNNs to capture spatial and temporal dependencies among stations respectively. \cite{chai2018bike} considered heterogeneous spatial relationships across stations using a graph convolutional network (GCN) model. A multi-relational GNN was developed by \cite{liang2022cross} to leverage spatial information from multi-modal data. However, all these methods are conditioned on large-scale historical data for model training and are not applicable to situations when the transportation systems are expanding from time to time.

Though limited, several recent works have shed light on spatiotemporal prediction in a data-scarce transportation system by transferring knowledge from data-rich transportation systems. For example, \cite{wang2018crowd} proposed a cross-city transfer learning algorithm for demand prediction by linking similar regions in source and target cities. A meta-learning approach was developed by \cite{yao2019learning} to transfer the model learned in multiple data-sufficient cities to cities with only a few days of historical transaction records. A domain-adversarial graph learning technique was introduced in \cite{tang2022domain} for short-term traffic prediction across cities. Although these research demonstrate the potential ability of deep learning models to generalize to new scenarios, they still focus on short-term demand prediction relying on sequential/temporal dependencies, rather than long-term demand forecasting to account for potential system changes and support transportation planning. For the latter, the spatial dependencies on the underlying built environment and across different stations are more fundamental and generalizable to TG-BSSE, and thus are the focus of this study.

\subsection{Enhancing Theory-based Models with Deep Neural Networks} \label{literature:synergy}

In the transportation field, data-driven and theory-based models are usually regarded as disparate. However, these two types of methods can be complementary: data-driven models usually demonstrate better prediction performance in data-rich environments, while theory-based methods (e.g., gravity model) are more advantageous in terms of generalizability and interpretability. Leveraging their complementary natures, several recent studies have explored the potential to combine DNNs and theory-based models. A theory-based residual neural network was introduced in \cite{wang2021theory} for choice analysis, which links DNNs with discrete choice models based on their shared utility interpretation. \cite{simini2021deep} interpreted the classic gravity model as a shallow neural network with restricted variables as input, and proposed Deep Gravity to generate flow probabilities between origin-destination (OD) pairs. \cite{zhu2022spatial} provided a comparative analysis of GCNs and linear spatial regression models, and demonstrated that the former can achieve a better performance in spatial imputation tasks. To further bridge the methodological gap between spatial regression and GNNs, this study will demonstrate that a linear spatial regression model can be regarded as a shallow neural network and our proposed GNN model is essentially a generalized spatial regression model. 

\section{Methodology} \label{sec:method}

In this section, a few important definitions and the problem formulation of our research are first introduced in Section~\ref{method:problem statement}. Next, we present a spatial GNN approach (Spatial-MGAT) to predict potential demand for BSS expansion in Section~\ref{method:architecture}. In Section~\ref{method:SRGNN}, we further elaborate on the relationship between our proposed GNN approach and classic spatial regression models.

\subsection{Problem Statement} \label{method:problem statement}

\textit{Definition 1 (BSS Station Demand):} For each BSS station, we aim to predict its average number of daily outflow (i.e., departure) and inflow (i.e., arrival) trips in different months. Such demand information is critical for strategic system planning decisions such as the choice of the station site and capacity, for which more detailed temporal (e.g., hourly) demand patterns are less important \citep{kouincorporating}. Since BSS demand varies greatly by seasons, with higher demand in summer months and lower demand in the winter (more details in Section~\ref{exp:data}), we choose to make monthly predictions to reflect such seasonal variability. The average daily bike outflow $y_{i,m}^{out}$ and inflow $y_{i,m}^{in}$ at station $i$ in a month $m$ is computed as:
\begin{equation}
\begin{array}{lcl}
    y_{i,m}^{out} = c_{i,m}^{out} / n_{i,m}, \\
    y_{i,m}^{in} = c_{i,m}^{in} / n_{i,m},
\end{array}
\end{equation}
where $c_{i,m}^{out}$ and $c_{i,m}^{in}$ are the number of departure and arrival trips at station $i$ in month $m$ respectively, $n_{i,m}$ is the number of days that station $i$ is active (i.e., with at least one departure or arrival trip) in month $m$. 

\textit{Definition 2 (Localized Station Network):} For a station $i$, to predict its demand in month $m$, we define a localized graph centered on station $i$ to capture its spatial dependencies with other related BSS stations, denoted as $G_{i,m}=(V_{i,m}, A_{i,m})$, where $V_{i,m}$ is a set of nodes (i.e., BSS stations) connected with node $i$ in month $m$, and $A_{i,m} \in \mathbb{R}^{|V_{i,m}| \times |V_{i,m}|}$ is a weighted adjacency matrix representing the dependencies between each pair of nodes in $V_{i,m}$. Note that both $V_{i,m}$ and $A_{i,m}$ can change over time due to BSS expansion.

\textit{Problem (Trip Generation for BSS Expansion):} This research aims to predict the number of trips originating and destined for each BSS station in an expanding system (with new stations added over time). Specifically, given the built environment features of a station $i$ in month $m$, denoted as $x_{i,m}$, its localized graph $G_{i,m}$, as well as the built environment features of other related stations in $G_{i,m}$, denoted as $XG_{i,m} = \{x_{j,m},\forall j \in V_{i,m}, j \neq i\}$, we learn a mapping function $F(\ast)$ to predict the inflow and outflow demand of station $i$ in month $m$, represented as $y_{i,m} = [y_{i,m}^{out}; y_{i,m}^{in}], y_{i,m} \in \mathbb{R}^2$:
\begin{equation}
y_{i,m}=F(x_{i,m}, G_{i,m}, XG_{i,m}).
\end{equation}

\subsection{Model Architecture}\label{method:architecture}

\begin{figure}[ht!]
  \centering
  \includegraphics[width=\linewidth]{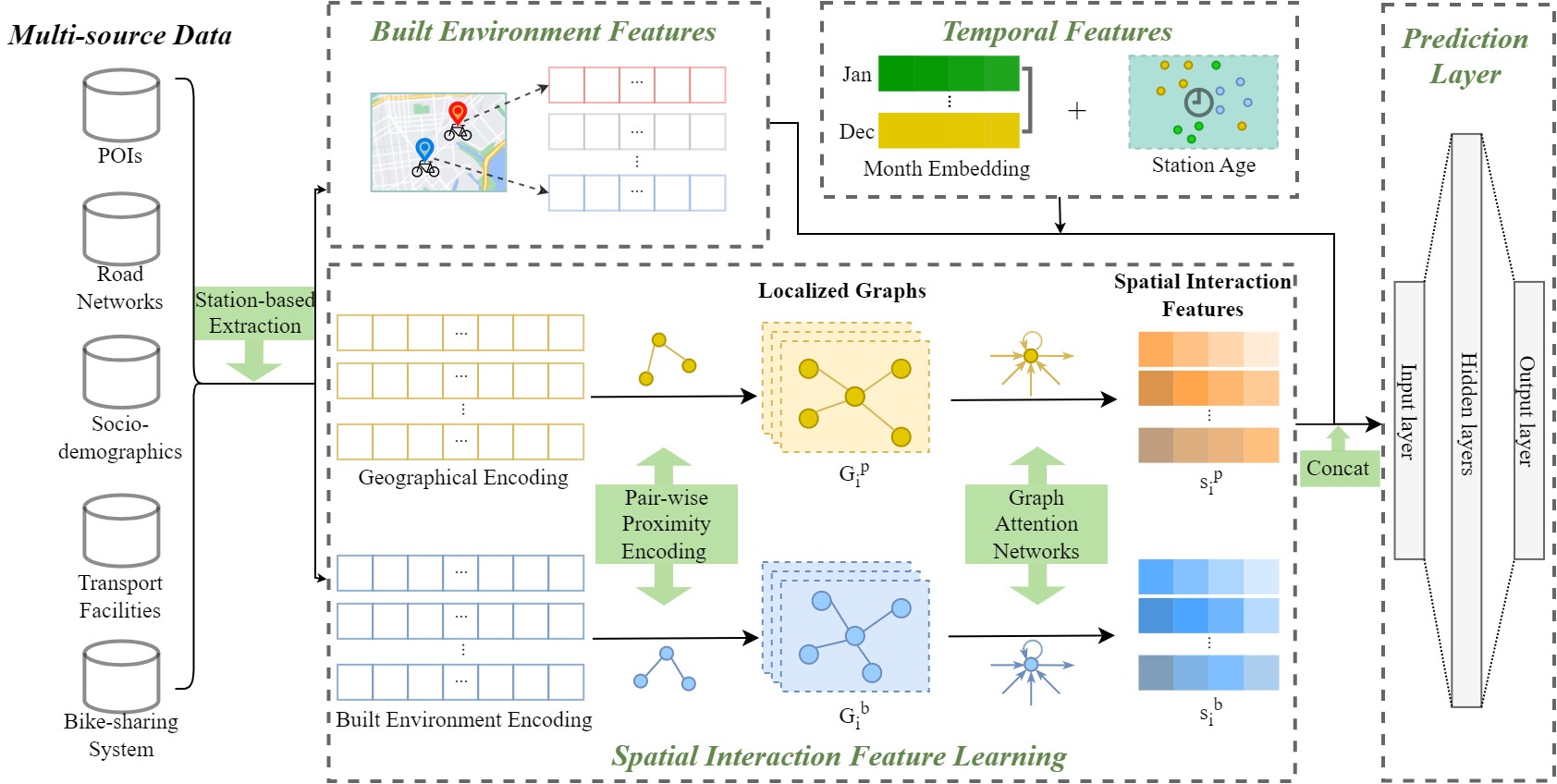}
  \caption{The architecture of Spatial-MGAT}
  \label{fig:architecture}
\end{figure}

In this section, we introduce our modeling framework for TG-BSSE based on multi-source urban data. The overview of the framework is shown in Fig.~\ref{fig:architecture}, which consists of four parts. First, we extract various geographic and demographic features to depict the built environment for BSS stations. Second, to capture spatial interactions between stations, we construct two localized graphs based on geographical proximity and built environment similarity respectively, and summarize the features of connected stations into spatial interaction feature vectors. Third, additional features are included to consider temporal information such as the month and station age. Finally, the aforementioned built environment, spatial interaction and temporal features are fed into an output layer to generate the number of inflow and outflow trips for the target station. 

\subsubsection{Built Environment Feature Extraction} \label{method:built_env}

Based on the existing literature, we extract a diverse set of built environment features for each BSS station, including the following categories:

\begin{itemize}[noitemsep]
    \item POI density (10 features): the number of POIs within a radius of 500m for each possible category, i.e., residential, educational, cultural, recreational, commercial, religious, transportation, government, health, and social services.
    \item Socio-demographics (11 features): demographic features of the census tract for the station, including population and housing unit density, proportion of population in households, proportion of people under 18, average household size, total housing units, proportion of occupied housing units, and proportion of different races (i.e., hispanic, white, asian and black).
    \item Road network (16 features): the total number and length of different road networks by levels, including motorway, trunk, primary, secondary, tertiary, unclassified and residential, the length of bike lanes, as well as the number of junctions within a radius of 500m.
    \item Transportation facilities (2 features): the distance to the nearest subway station and the number of subway stations within a radius of 500m.
    \item BSS network design (4 features): the number of BSS stations within 0-500m, 500-1,000m and 1,000-5,000m travel distance respectively, average travel distance to the other BSS stations.
\end{itemize}

After getting the above features, each BSS station $i$ is related with a 43-dimensional vector $x_i$ to describe its surrounding built environment. It is worth noting that $x_i$ can change over time. For example, as the system network is evolving, the number of BSS stations in its neighborhood can dynamically change. For a detailed description of data sources of the aforementioned variables, please refer to Section~\ref{exp:data}.

\subsubsection{Localized Graph Construction}

For trip generation given a target BSS station, it is important to consider not only the effect of its local built environment, but also its interactions with other stations. To capture spatial correlations among stations with close geographical distances or similar built environment, we encode two types of spatial dependencies for each station pair:

\textit{Geographical Proximity:} According to the First Law of Geography, station pairs that are geographically adjacent are more likely to be strongly correlated than distant ones. We define a geographical proximity weight between each station pair $i$ and $j$ as:
\begin{equation}
a_{ij}^p =\exp(-(\frac{d_{ij}}{\sigma_d})^2),
\end{equation}
where $a_{ij}^p$ is the geographical proximity weight between stations $i$ and $j$, $d_{ij}$ is the geographic distance between $i$ and $j$, and $\sigma_d$ is the standard deviation of distances. Based on the geographical proximity weights computed above, we construct a localized graph for a target BSS station $i$ consisting of its $k$ nearest neighbors, denoted as $G_{i}^p$. $G_{i}^p$ can be updated as new stations are added to the system, since the $k$-nearest neighbors of the same station can dynamically change. 

\textit{Built Environment Similarity}: Previous research has shown that in addition to geographically nearby stations, incorporating features of stations with similar built environment or functionalities can also enhance the prediction performance \citep{zhou2021modeling}. We define a built environment similarity weight between each station pair to measure such ``semantic'' relationships:
\begin{equation}
a_{ij}^b =\exp(-(\frac{dist(x_i, x_j)}{\sigma_b})^2),
\end{equation}
where $a_{ij}^b$ is the weight of built environment similarity between stations $i$ and $j$, $x_i, x_j$ are the built environment feature vectors of station $i$ and $j$ as defined in Section~\ref{method:built_env}, $dist(\ast)$ is an Euclidean distance function to compute the distance between $x_i$ and $x_j$, and $\sigma_b$ is the standard deviation of all the vector distances. For a target station $i$, the $k$ BSS stations that have the most similar built environment features are selected, resulting in a localized graph denoted as $G_{i}^b$. Note that $G_{i}^b$ can also change over time due to newly added stations as well as the variation of $x_i$ and $x_j$.

\subsubsection{Spatial Interaction Feature Learning}

Given the pre-defined localized graphs, we generate two spatial interaction feature vectors for a target station $i$ based on $G_{i}^p$ and $G_{i}^b$, denoted as $s_{i}^p$ and $s_{i}^b$, respectively. The spatial interaction feature vector is computed as a weighted sum of the built environment features of its connected stations and the weight is learnt through an attention mechanism. Taking $G_{i}^p$ as an example, we compute its corresponding spatial interaction feature vector through the following steps:

\textit{Step 1}: Encode the built environment features of station $i$ as well as its connected stations in the localized graph to a latent vector using a shared linear layer: 
\begin{equation}
\label{eq:adj_feat}
    h_j = W_h x_j + b_h, \forall j \in V_{i}^p,
\end{equation}
where $h_j \in \mathbb{R}^{d_h}$ is a $d_h$-dimensional encoded vector for station $j$, $W_h \in \mathbb{R}^{43 \times d_h}$ and $b_h \in \mathbb{R}^{d_h}$ are the learned model parameters for feature encoding.

\textit{Step 2}: Compute the importance score of each connected station $j \in  ,V_{i}^p j \neq i$ to station $i$ using a shared feed-forward network applied to every station pair:
\begin{equation}
\label{eq:attn}
\begin{array}{lcl}
z_{ij} = ReLU(W_{s,1}[h_i; h_j] + b_{s,1}),\\
s_{ij} = LeakyReLU(W_{s,2}z_{ij} + b_{s,2}),
\end{array}
\end{equation}
where $s_{ij}\in \mathbb{R}$ is the importance score of station $j$ to station $i$, $W_{s,1} \in \mathbb{R}^{2 d_h \times d_z}, W_{s,2} \in \mathbb{R}^{d_z \times 1}, b_{s,1} \in \mathbb{R}^{d_z}, b_{s,2} \in \mathbb{R}^1$ are parameters to be learned, $d_z$ is the dimension of the hidden vector $z_{ij}$. 

\textit{Step 3}: The attention weight of connected stations to station $i$ is then computed by normalizing the importance score using a softmax function: 
\begin{equation}
\label{eq:softmax}
\epsilon_{ij}=\frac{\exp(s_{ij})}{\sum_{j \in V_{i}^p, j \neq i} \exp(s_{ij})},
\end{equation}
where $\epsilon_{ij} \in \mathbb{R}$ is the attention weight of connected station $j$ to target station $i$. 

\textit{Step 4}: The spatial interaction vector $s_{i}^p \in \mathbb{R}^{d_h}$ is computed as a weighted sum of the features of connected stations:
\begin{equation}
s_{i}^p = \sum_{j \in V_{i}^p, j \neq i} \epsilon_{ij} h_{j}.
\end{equation}

Using the same method, we compute $s_{i}^b \in \mathbb{R}^{d_h}$ based on $G_{i}^b$. The spatial interaction vectors will be further used as input features for the prediction layer as introduced later.

\subsubsection{Temporal Feature Learning} \label{method:temp}
While the proposed model focuses on learning spatial dependencies across stations, the effect of relevant temporal factors should not be overlooked. Specifically, we include two additional temporal features in our model, i.e., month and station age (i.e., the number of months since a target station opens). The month feature is originally a categorical variable and a simple way to encode it as model input is to use one-hot vectors, which, however, might not capture seasonal fluctuations across different months. Instead, we use embedding techniques to map each month to a representation vector. Specifically, a parameter matrix $W_m \in \mathbb{R}^{12 \times d_m}$ is learnt in our model, with each row representing each month as a $d_m$-dimensional vector.

\subsubsection{Prediction Layer}
The prediction layer aims to generate the final trip prediction based on the aforementioned built environment, spatial interaction and temporal features. Specifically, for a target station $i$ in month $m$, the prediction layer takes the following form:  
\begin{equation}
\label{eq:prediction}
\begin{array}{lcl}
    z_{o,1} = ReLU(W_{o,1}([x_{i,m}; s_{i,m}^p; s_{i,m}^b; t_{i,m}]) + b_{o,1}), \\
    z_{o,2} = sigmoid(W_{o,2} z_{o,1} + b_{o,2}), \\
    \hat{y}_{i,m} = W_{o,3} z_{o,2} + b_{o,3},
\end{array}
\end{equation}
where $\hat{y}_{i,m} \in \mathbb{R}^2$ is the predicted demand of station $i$ in month $m$, $t_{i,m} \in \mathbb{R}^{d_m + 1}$ is the temporal feature vector of station $i$ in month $m$, $W_{o,1} \in \mathbb{R} ^ {(43 + 2d_h + d_m + 1) \times d_{o}^1}, W_{o,2} \in \mathbb{R} ^ {d_{o}^1 \times d_{o}^2}, W_{o,3} \in \mathbb{R} ^ {d_{o}^2 \times 2}$ are parameter matrices and $b_{o,1} \in \mathbb{R}^{d_{o}^1}, b_{o,2} \in \mathbb{R}^{d_{o}^2}, b_{o,3} \in \mathbb{R}^2$ are model bias terms, and $d_{o}^1,d_{o}^2$ are dimensions of hidden vectors $z_{o,1}$ and $z_{o,2}$, respectively.

The model is trained to minimize the sum of squared errors between predicted and real demand values across BSS stations over time:
\begin{equation}
\label{eq:loss}
L_{\theta} = \sum_{m=1}^{M} \sum_{i=1}^{N_m}(\hat{y}_{i,m} - y_{i,m})^2,
\end{equation}
where $M$ is the number of months in the training data and $N_m$ denotes the number of active stations in month $m$.

\subsection{Discussion: Synergy between Spatial Regression and GNNs}\label{method:SRGNN}

In the previous section, we introduce our model architecture as a GNN approach. In this section, we provide an alternative way to understand our model as a generalized version of a conventional spatial regression model named the spatial lag of X (SLX) \citep{elhorst2017slx}. Specifically, we first introduce the original SLX model in Section~\ref{method: slx} and then elaborate on how Spatial-MGAT extends SLX in Section~\ref{method: nonlinear_slx}. 

\subsubsection{SLX Model}\label{method: slx}

SLX is one of the first and most straightforward spatial regression models, which incorporates the average value of explanatory variables from surrounding locations as model input \citep{elhorst2017slx}. Mathematically, it is given as:
\begin{equation}
\begin{array}{lcl}
\hat{y}_i = f([x_i; s_i^p]), \\
s_i^p = \sum_{j \in V_{i}^p, j \neq i}{w_{ij}^p h_j^p},
\end{array}
\end{equation}
where $\hat{y}_i$ and $x_i$ is the dependent variable and explanatory variables of location $i$ respectively, $s_i^p$ is a spatial lag term that captures the average spatial information from neighborhood regions of location $i$. $V_{i}^p$ denotes a set of locations adjacent to $i$, $w_{ij}^p$ is a pre-defined adjacency weight between locations $i$ and $j$, and $h_j^p$ is a subset of location $j$'s explanatory variables that are thought to be relevant to spatial dependencies. $f(\star)$ is a function to be learned assuming linear relationships between input and dependent variables. 

In this study, the dependent variable is the number of trips generated for each BSS station in different months, and the explanatory variables include built environment features (see Section~\ref{method:built_env}) and temporal information (see Section~\ref{method:temp}). Therefore, the SLX model for TG-BSSE can be expressed as:
\begin{equation}
\begin{array}{lcl}
\hat{y}_{i, m} = f([x_{i,m}; s_{i,m}^p; t_{i, m}]), \\
s_{i,m}^p = \sum_{j \in V_{i,m}^p, j \neq i}{w_{ij,m}^p h_{j,m}^p},
\end{array}
\end{equation}
where $\hat{y}_{i, m}$ is the predicted demand of station $i$ in month $m$, $x_{i,m}$ and $s_{i,m}^p$ are the built environment features and spatial lag term respectively, $t_{i, m}$ is a temporal feature vector with the month feature encoded as a one-hot vector. 

Since the spatial lag term is a pre-defined spatial transformation of the explanatory variables of nearby locations, the linear function $f(\star)$ can be easily estimated using ordinary least squares (OLS). Alternatively, the model may be regarded as a single-layer linear neural network without any activation function, which can be estimated by minimizing the squared loss defined in Eq.~\eqref{eq:loss}. This allows us to naturally extend the SLX model with GNNs.

\subsubsection{Spatial-MGAT as a Generalized SLX Model} \label{method: nonlinear_slx}

As introduced in the previous section, SLX can be regarded as a linear neural network and naturally extended by adding hidden layers and nonlinear activation functions. Specifically, our proposed Spatial-MGAT can be regarded as a generalized SLX, with modifications in the several aspects. First, to capture multiple types of spatial dependencies, we incorporate an additional spatial lag term $s_i^b$ to capture the average spatial information from BSS stations with similar built environment. Mathematically, this can be expressed as:
\begin{equation}
\begin{array}{lcl}
\hat{y}_{i, m} = f([x_{i,m}; s_{i,m}^p; s_{i,m}^b; t_{i, m}]), \\
s_{i,m}^p = \sum_{j \in V_{i,m}^p, j \neq i}{w_{ij,m}^p h_{j,m}^p}, \\
s_{i,m}^b = \sum_{j \in V_{i,m}^b, j \neq i}{w_{ij,m}^b h_{j,m}^b}, \\
\end{array}
\end{equation}
where $V_{i,m}^b$ denotes a set of BSS stations that have similar built environment features with station $i$ in month $m$. Second, instead of manually determining the subset of explanatory features from neighborhood stations, we use a linear transformation layer to learn the representation feature vectors of connected stations, denoted as $h_{j,m}^p$ and $h_{j,m}^b$ (see Eq.~\eqref{eq:adj_feat}). Third, rather than using pre-defined adjacency weights, the model learns the adjacency weights $w_{ij,m}^p$ and $w_{ij,m}^b$ with attention mechanisms (see Eq.~\eqref{eq:attn} and Eq.~\eqref{eq:softmax}). Fourth, categorical variables, such as the month, can be represented as lower-dimensional embedding vectors, instead of one-hot vectors, to be learned simultaneously in model training. Finally, the prediction function $f(\star)$ is replaced with a feed-forward network as expressed in Eq.~\eqref{eq:prediction}. As we will show in Section~\ref{res: tra_method}, with the aforementioned simple modifications, our proposed model can significantly outperform the original SLX model. This demonstrates the potential of using DNNs to enhance the performance of traditional econometric models.

\section{Results} \label{sec:experiments}

\subsection{Data Description} \label{exp:data}
To validate the performance of our proposed model, we use the Citi Bike system in New York City (NYC) as a case study. The Citi Bike data \footnote{https://ride.citibikenyc.com/system-data} provides the start and end station and time of each trip over time. Citi Bike was first launched in July 2013 and has undergone massive expansions in the years since. Our preliminary analysis shows that the usage pattern has become quite different since 2020 (likely due to the COVID-19 pandemic). To circumvent the impact of COVID-19, the data from 2013-07 to 2019-12 is used for empirical analysis. Prior to model evaluation, an overview of the data is presented to offer a better understanding of the system expansion process and the spatiotemporal demand distribution of BSS stations.

\begin{figure}[!htb]
    \centering
    \includegraphics[width=0.9\linewidth]{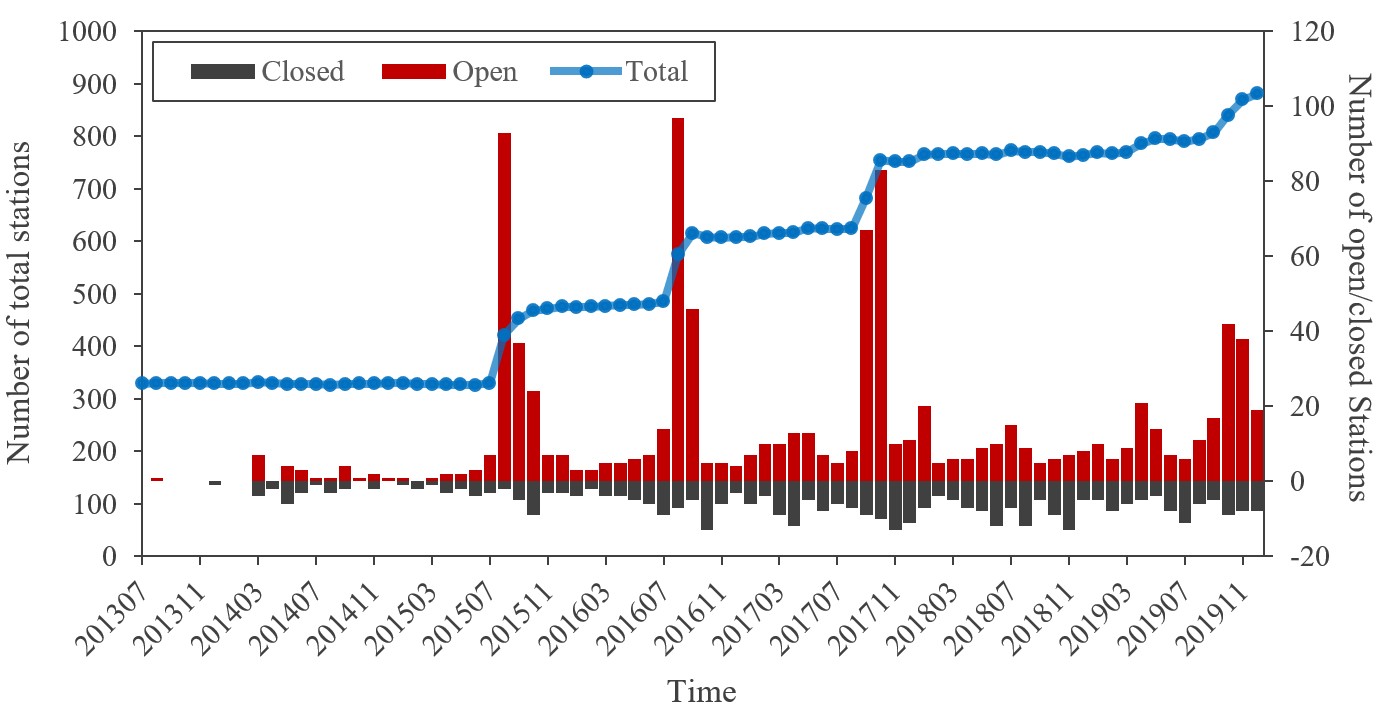}
    \caption{BSS expansion process in NYC over time}
    \label{fig:temporal_expansion}
\end{figure}

\begin{figure}[!htb]
    \centering
    \includegraphics[width=0.5\linewidth]{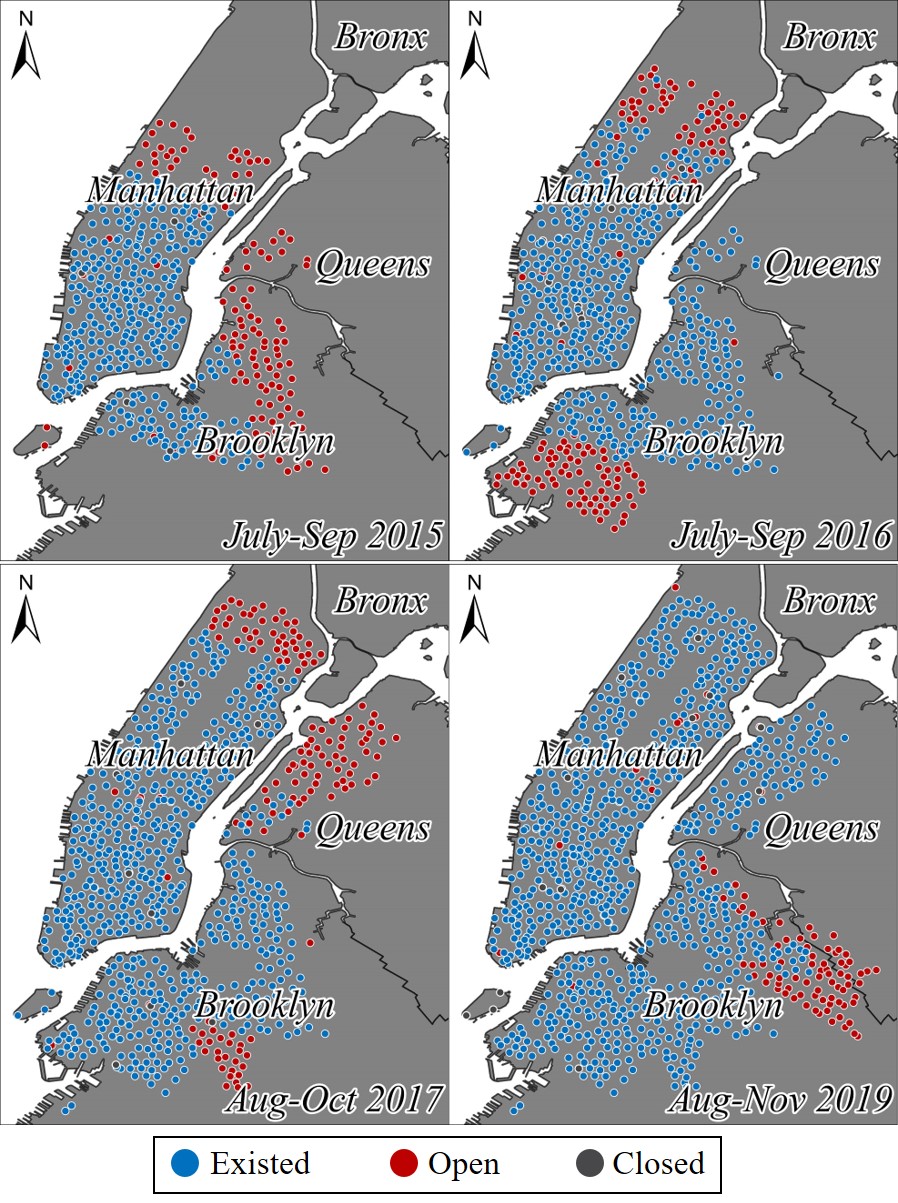}
    \caption{Spatial distribution of BSS stations before and after mass development periods}
    \label{fig:spatial_expansion}
\end{figure}

Fig.~\ref{fig:temporal_expansion} presents the expansion process of NYC BSS during our study period. The blue line represents the total number of BSS stations, and the red and grey bars indicate the number of newly opened and closed stations\footnote{We consider a station as closed if it observes no trip for at least a month.}. It can be clearly seen that from 2013-07 to 2019-12, there are 4 mass expansion periods, i.e., 2015-08, 2016-08, 2017-09, and 2019-09. Outside of these periods, the system is dynamically evolving with newly opened and closed stations simultaneously, while the total number of stations keeps almost unchanged. The spatial distribution of BSS stations before and after the 4 mass expansion periods is shown in Fig.~\ref{fig:spatial_expansion}. Blue dots represent unchanged BSS stations before and after expansion, while red and grey dots represent newly opened and closed bike stations after expansion. Before July 2015, BSS stations in NYC were only distributed in Downtown and Midtown Manhattan, as well as in the west part of Brooklyn (e.g., Downtown Brooklyn and Williamsburg). From 2015 to 2016, the system was extended to Upper Manhattan, and the north and southwest sides of Brooklyn. BSS stations in Manhattan were expanded further north in 2017, and new stations were built in the west part of Queens (e.g., Long Island City and Astoria) and southeast side of Brooklyn. In 2019, BSS expansion in Manhattan came to a near halt, with new development concentrated in the east side of Brooklyn. In comparison, the removal of stations is almost negligible. 

\begin{figure}[!htb]
    \centering
    \includegraphics[width=0.85\linewidth]{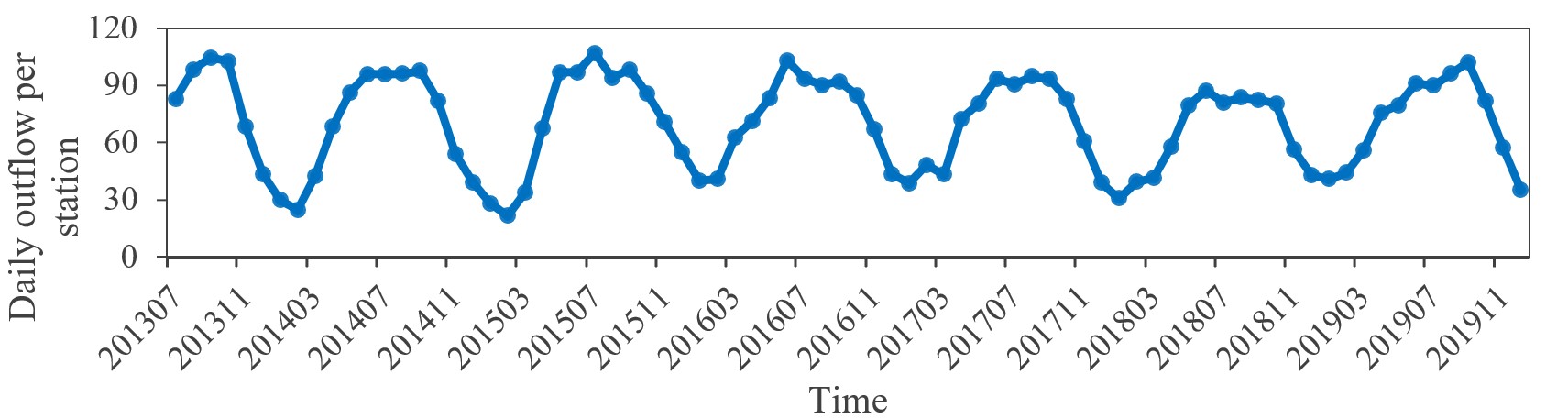}
    \caption{Seasonal pattern of average BSS station demand}
    \label{fig:temporal_demand}
\end{figure}

\begin{figure}[!htb]
    \centering
    \includegraphics[width=0.8\linewidth]{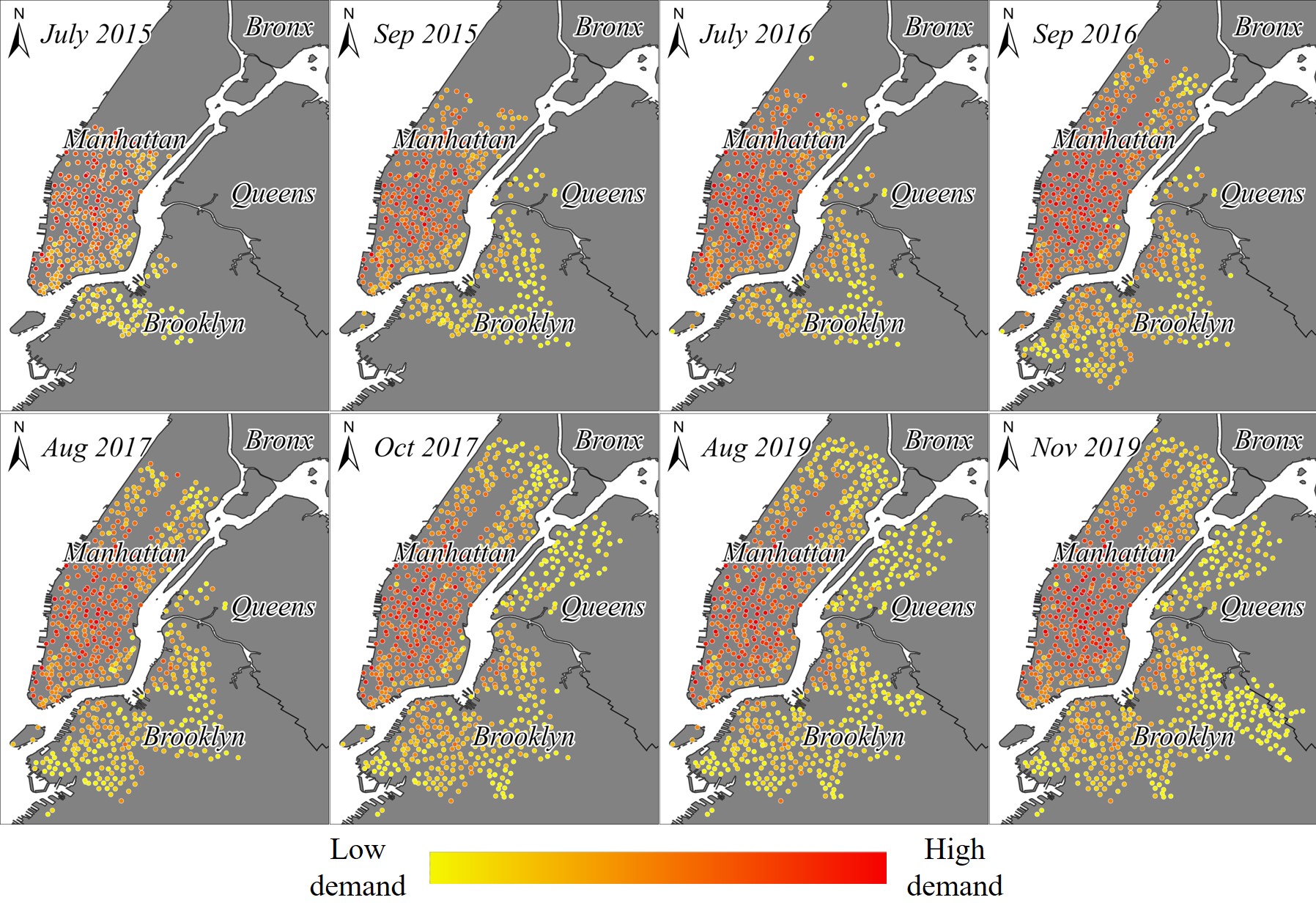}
    \caption{Spatial pattern of BSS station demand in NYC over time}
    \label{fig:spatial_demand}
\end{figure}

The temporal pattern of daily average outflow trips per station in different months is presented in Fig.~\ref{fig:temporal_demand}. A clear seasonal pattern can be observed, with higher ridership in the summer and autumn and lower ridership in the winter and spring. This can be explained by the weather fluctuations, as low temperatures and higher chances of snow can deter people from cycling outdoors. It is worth noting that, while the total BSS demand grows over time (as a result of system expansion), the average trip generation per station does not change much. This is despite the fact that stations added in later years are typically in lower-density neighborhoods and generally have lower demand.
The spatial distributions of BSS trips in several selected months are presented in Fig.~\ref{fig:spatial_demand}. It shows a pattern of gradual decay from the downtown area of Manhattan to its proximity, suggesting that Downtown and Midtown Manhattan have the highest demand for bike sharing, while the demand in other areas are relatively lower.

In addition to BSS trip data, several open-source supplementary data are used for built environment feature extraction:

\begin{itemize} [noitemsep]
    \item POIs: The POI data is obtained from NYC Open data\footnote{https://opendata.cityofnewyork.us/} which consists of 20,558 POI points from 13 facility categories. We use 10 of them as listed in Section~\ref{method:built_env}.
    \item Socio-demographics: The data comes from NYC Population Factfinder\footnote{https://popfactfinder.planning.nyc.gov/}, which provides Decennial Census data in 2010 for different census tracts in NYC.
    \item Road Network: The road network data is downloaded from OpenStreetMap\footnote{https://www.openstreetmap.org}, comprising
    91,868 roads and 55,314 junctions in NYC. The downloaded data associates each road link with a road level and we consider 7 of them for feature construction as listed in Section~\ref{method:built_env}. The bike lane data is obtained from NYC Open Data, which provides the spatial distribution and opening dates of different bike lanes in NYC. 
    \item Transportation facilities: We obtain the distribution of subway stations from NYC Open Data, with 218 subway stations in total.
\end{itemize}

\subsection{Experiment Design}

To reflect the realistic effect of system expansion over time, we use BSS trip data from 2013-07 to 2017-08 for model training and validation, and data from 2017-09 to 2019-12 for model testing. During 2013-07 to 2017-08, there are 21,827 station-month observations across 734 stations. We randomly select 80\% for model training and the rest 20\% for model validation, resulting in 17,462 and 4,365 training and validation samples respectively. During the testing period, there are 1,012 stations in total with 21,808 station-month observations. Among the 1,012 stations, 645 stations exist in the training set with 16,446 station-month observations and 367 stations are newly added (i.e., unseen in the training and validation set) with 5,362 station-month observations. We use the monthly observations of both existing and newly added stations as test set and evaluate our model performance on them separately. 

To facilitate model training, min-max normalization is applied to the demand data as well as the input built environment variables before feeding them into the model. For deep learning models, the number of training epochs $E$ is set as 200 and we use early stopping of 10 epochs on the validation set to prevent overfitting. The models are trained using Adam Optimizer with a learning rate of 0.002, a batch size of 32 and L2 regularization with a weight decay equal to 1e-5. The hyperparameters of our proposed model are set as follows: the number of neighbors $k$=5,  the dimension of spatial interaction feature vectors $d_h$=8, the hidden dimension for attention weight learning $d_z$=16, the dimension of month embedding vectors $d_m$=12, and the hidden dimensions of the prediction layer $d_o^1$=32, $d_o^2$=16. We repeat experiments for each model 10 times and report the average performance. The model performance is evaluated using 3 metrics computed on the test set: root mean square error (RMSE), mean absolute error (MAE) and the coefficient of determination ($R^2$).

\subsection{Comparison with regression and machine learning models}\label{res: tra_method}

Previous research for TG-BSSE typically used regression or tree-based ensemble machine learning models to capture the relationship between BSS demand and urban built environment. In this subsection, we compare the performance of our proposed model with the following regression and machine learning baselines:

\begin{itemize} [noitemsep]
    \item \textbf{Linear Regression} \citep{singhvi2015predicting}: a regression model which assumes linear relationships between BSS station demand and input variables. In our implementation, its input variables include built environment (Section~\ref{method:built_env}) and temporal features (Section~\ref{method:temp}), with the month feature encoded as a one-hot vector. 
    \item \textbf{Spatial Regression} \citep{faghih2016incorporating}: We use the original SLX model introduced in Section~\ref{method: slx} as an example of spatial regression models, which incorporates spatial dependencies between geographically close stations by including built environment features from nearby BSS stations as additional model input.
    \item \textbf{XGBoost} \citep{kouincorporating}: a tree-based ensemble machine learning method based on gradient boosting decision trees. We use the same input variables as those of linear regression in our implementation.
    \item \textbf{Function Zone} \citep{liu2017functional}: a hierarchical model which first clusters stations into functional zones, and then predict bike sharing trip generation from functional zone level to station level using Random Forest and Ridge Regression respectively. We replace Ridge Regression with XGBoost, which can achieve better performance in our dataset.
\end{itemize}

The results of different models are presented in Table~\ref{table:baseline_model}. It is found that all models result in larger RMSE and MAE for existing stations than new stations. This is reasonable as existing stations are mainly distributed in downtown regions (see Figure~\ref{fig:spatial_demand}) and are associated with higher demand. Meanwhile, the $R^2$ for existing stations is higher than that for new stations using all methods. This can be explained that the models are trained with historical observations of existing stations. Compared with the baseline models, our approach achieves significantly superior performance regarding all evaluation metrics for both new and existing stations. This is likely because our method leverages deep learning techniques to capture the nonlinear relationship between BSS demand and input variables, and graph learning approaches to model complex spatial interactions across BSS stations. 

Among baseline models, the poor performance of regression approaches is likely due to their oversimplified linear assumptions. Leveraging spatial information from nearby BSS stations, spatial regression performs better than linear regression for existing stations. However, it does not provide notably better prediction for newly planned stations in our experiments. This might be because spatial interactions among BSS stations are quite complicated and cannot be effectively captured using simple linear models. Benefiting from the ability of machine learning models to capture relationships from data, XGBoost improves the prediction performance by a large margin compared to regression models. The advantage of Function Zone to XGBoost is minimal in our case, which might be because our implementation does not include taxi transaction record data used in the original paper. Compared with XGBoost, Spatial-MGAT can further reduce the prediction error, with RMSE improvement of 22.6\% and 32.1\% for newly planned and existing stations respectively. As introduced in Section~\ref{method:SRGNN}, our method can be regarded as a generalized spatial regression model. Compared with the linear spatial regression model, Spatial-MGAT can reduce RMSE by 71.5\% and 40.3\% for new and existing stations. This suggests that DNNs, with a proper network design, can greatly enhance the model performance of classical econometric models, and the improvement can generalize well to unseen observations (e.g., new BSS stations). 

\begin{table}[ht!]
  \centering \footnotesize
  \caption{Performance comparison among regression and machine learning models}
    \begin{tabular}{ccccccc}
    \toprule
    \multirow{2}{*}{Models} & \multicolumn{3}{c}{New stations} & \multicolumn{3}{c}{Existing Stations} \\
    & RMSE & MAE & $R^2$ & RMSE & MAE & $R^2$\\
    \midrule
    Linear Regression & 33.646 & 26.185 & 0.454 & 49.369 & 36.692 & 0.555 \\
    Spatial Regression & 34.963 & 26.812 & 0.467 & 46.662 & 34.606 & 0.596  \\
    XGBoost & 25.920 & 16.900 & 0.602 & 41.027 & 27.719 & 0.717 \\
    Function Zone & 26.343 & 16.882 & 0.576 & 42.311 & 29.004 & 0.655 \\
    Spatial-MGAT & \underline{\textit{20.392}} & \underline{\textit{12.599}}  & \underline{\textit{0.742}} & \underline{\textit{27.860}} & \underline{\textit{19.140}} &  \underline{\textit{0.852}}\\
    \bottomrule
    \end{tabular}
  \label{table:baseline_model}%
\end{table}%

\subsection{Comparison with deep learning variants}

As introduced in Section~\ref{literature:new}, existing deep learning approaches for TG-BSSE require historical demand data of nearby BSS stations as input and are formulated quite differently from our problem. In \ref{appendix:a}, we test their performance in real-world BSS expansion scenarios. It turns out that they may not work well due to data sparsity and demand distribution discrepancy issues. To the best of our knowledge, our study is among the first to apply deep learning techniques to for TG-BSSE based on urban built environment features. To properly evaluate the effectiveness of our model architecture and quantify the contribution of different model components, we design several variant models as listed below:

\begin{itemize} [noitemsep]
    \item \textbf{Feed Forward Network (FNN)} \citep{svozil1997introduction}: A general FNN consists of an input layer, an output layer and several hidden layers in between. For fair comparison, FNN takes the form of the prediction layer used in our model with features of the target station as input. It will be used as the deep learning baseline for model benchmarking.
    \item \textbf{Multi-graph Convolutional Network (Spatial-MGCN)} \citep{kipf2016semi}: Spatial-MGCN adopts a similar structure with Spatial-MGAT, with GAT replaced by GCN to model spatial interactions among stations. The main difference between GAT and GCN is that GAT adaptively learns attention weights to represent correlations between station pairs, while GCN uses pre-defined adjacency weights to capture spatial dependencies. Note that Spatial-MGCN can also be regarded as an extended version of SLX with pre-defined weights between neighborhoods.
    \item \textbf{Graph Attention Network based on geographical proximity (Spatial-PGAT)}: In our proposed model, two types of spatial dependencies are considered: geographical proximity and built environment similarity. This variant model only encodes spatial interactions between geographically nearby stations. Spatial-PGAT can also be seen as a generalized SLX with only the spatial lag term for geographical proximity.
    \item \textbf{Graph Attention Network based on built environment similarity (Spatial-BGAT)}: In this variant, with the dependency of geographical proximity ablated, the model makes prediction based on spatial interaction features from only stations with similar built environment characteristics. Similar, Spatial-BGAT can be regarded as a generalized SLX with only the spatial lag term for built environment similarity.
\end{itemize}

\begin{table}[ht!]
  \centering \footnotesize
  \caption{Performance comparison among various deep learning models}
    \begin{tabular}{ccccccc}
    \toprule
    \multirow{2}{*}{Models} & \multicolumn{3}{c}{New stations} & \multicolumn{3}{c}{Existing stations} \\
    & RMSE & MAE & $R^2$ & RMSE & MAE & $R^2$\\
    \midrule
    FNN & 23.673 & 14.528 & 0.659 & 29.492 & 20.123 & 0.834 \\
    Spatial-MGCN & 21.654 & 13.042 & 0.716 & 28.645 & 19.710 & 0.845  \\
    Spatial-PGAT & 20.965 & 12.822 & 0.729 & 28.334 & 19.544 & 0.848\\
    Spatial-BGAT & 21.924 & 13.757 & 0.704 & 29.448 & 19.894 & 0.834\\
    Spatial-MGAT & \underline{\textit{20.392}} & \underline{\textit{12.599}} & \underline{\textit{0.742}} & \underline{\textit{27.860}} & \underline{\textit{19.140}} & \underline{\textit{0.852}}\\
    \bottomrule
    \end{tabular}
  \label{table:dnn_baseline}%
\end{table}%

Table~\ref{table:dnn_baseline} displays the average performance of our proposed approach and the variant deep learning models over 10 independent runs. It is found that FNN can already achieve notably better performance than XGBoost, demonstrating the ability of deep architectures in capturing complex nonlinear relationships between BSS demand and input variables. Using GCNs to leverage spatial information from related BSS stations, Spatial-MGCN performs better than FNN by a large margin, suggesting the effectiveness of considering station interactions using graph learning approaches. Meanwhile, Spatial-MGAT can further reduce the prediction error compared with Spatial-MGCN, which is likely due to the use of attention mechanisms to adaptively learn correlation weights between stations. With either geographical proximity or built environment similarity graph ablated, the variant model performs worse than the original one. This verifies the importance of using multiple graphs to capture heterogeneous relationships between BSS stations. Between the two, geographic proximity has a greater impact on prediction results, which is reasonable as BSS stations are more likely to be influenced by other stations that are spatially close by. 

\begin{figure}[ht!]
    \centering
    \includegraphics[width=\linewidth]{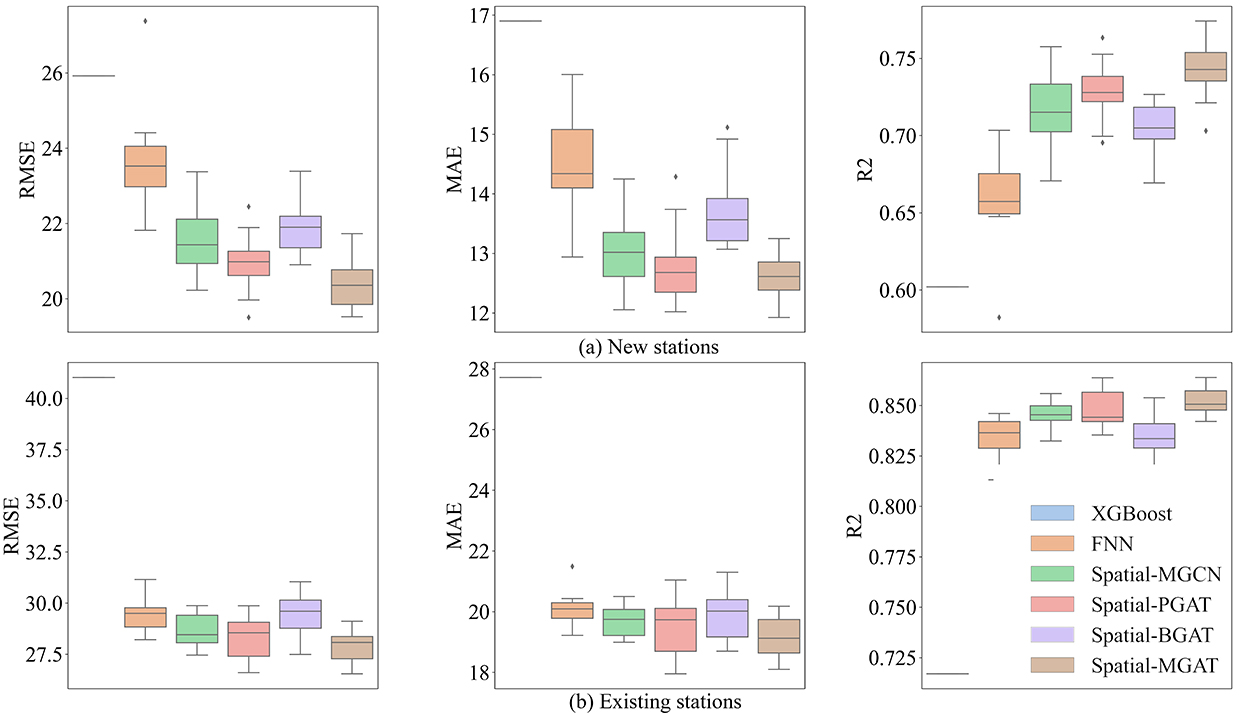}
    \caption{Comparison of model stability}
    \label{fig:box_plot}
\end{figure}

A major concern of DNNs is their instability in practical applications. To investigate this, we display the prediction results of all deep learning models in 10 independent experiments and use XGBoost as a benchmark in Figure~\ref{fig:box_plot}. It can be found that compared to FNN, graph learning models display relatively smaller variance. This suggests that incorporating the built environment features of connected stations can potentially improve model stability. In addition, although graph deep learning approaches generally have larger performance variance than XGBoost, they can perform significantly better than XGBoost for both newly planned and existing stations in all experiments. This demonstrates that GNNs can be a promising solution for the TG-BSSE problem with reasonable performance variability. 

\begin{figure}[ht!]
    \centering
    \includegraphics[width=\linewidth]{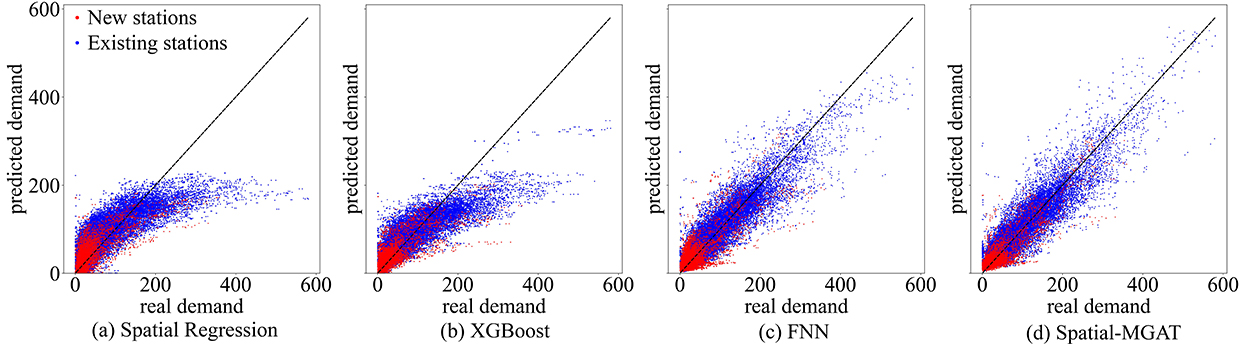}
    \caption{The relationship between true and predicted demand of several selected models}
    \label{fig:predict_true}
\end{figure}

To further provide an intuitive comparison of different model families, we plot the relationships between the true and predicted number of trips for all station-month observations in the test set based on several selected models in Figure~\ref{fig:predict_true}. Specifically, we use spatial regression as an example of regression models, XGBoost as an example of machine learning models, FNN as an example of non-graph deep learning models, and Spatial-MGAT as an example of graph learning models. 
It is apparent that Spatial-MGAT can provide a better model fit, especially for high-demand observations. In BSS or many other transportation systems, it is typical that a small number of high-demand stations serve a large proportion of passengers (i.e., the Matthew effect), and thus the ability to generate accurate predictions for these stations is essential for the efficiency of the whole system. In comparison, traditional models do not perform as well for such high-demand stations, as evidenced in Figure~\ref{fig:predict_true}(a-b), making them less practically useful. This demonstrates the value of using deep architectures and learning nonlinear relationships, especially when the data is unevenly distributed.

\subsection{Understanding the relationship between built environment and BSS demand}

In addition to trip generation for newly planned stations, understanding how the built environment would affect BSS demand is important to provide policy implications for BSS network design. In this section, we unravel the effects of different built environment features based on the model results using Shapley Additive exPlanations (SHAP). SHAP is an explainable AI technique whose main idea is to explain machine learning models using game theories \citep{lundberg_unified_2017}. With SHAP, each feature is assigned with an optimal Shapley value, which suggests how the presence or absence of a feature influences the model prediction result (i.e., BSS station demand in our case). 

\begin{figure}[ht!]
    \centering
    \includegraphics[width=0.85\linewidth]{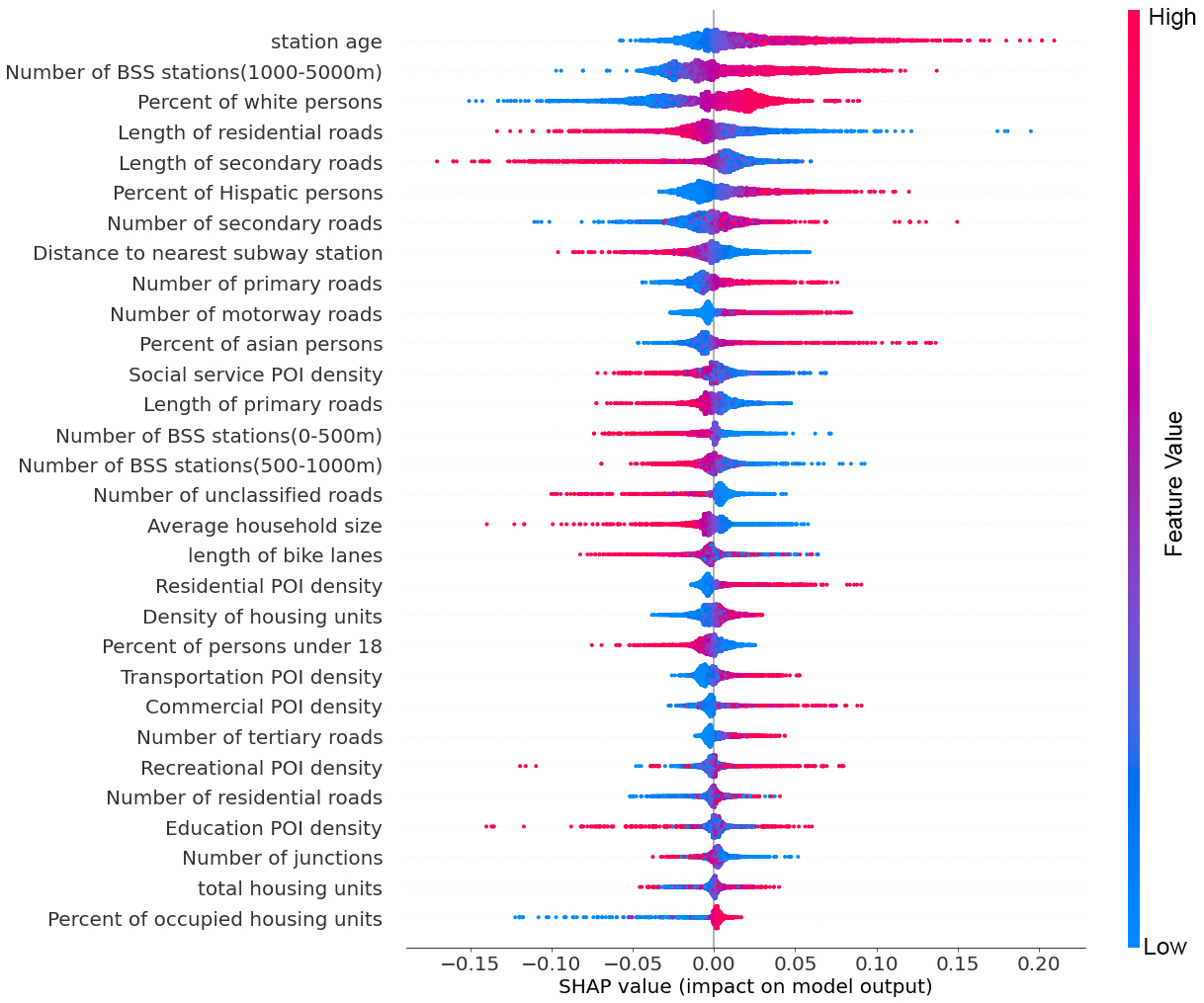}
    \caption{Distribution of Shapley values for the top 30 features in Spatial-MGAT}
    \label{fig:shap}
\end{figure}

Fig.~\ref{fig:shap} shows the distribution of SHAP values of the 30 most influential features in our proposed model. It is found that the station age plays the most important role. The longer a station has been in the system, the higher its demand is. This is as expected, since a newly added station is generally less well known to users, resulting in relatively low usage in the beginning. The number of BSS stations in the neighborhood also plays an important role: generally, the demand for a BSS station can increase with more stations in the 1000-5000m distance range, but decrease with more stations within 1000m travel distance. This suggests that stations that are too close to each other can compete for users, while stations in a medium distance can complement each other and attract more users. Another influential factor is the distance to the nearest subway station: BSS stations that are closer to a subway station are usually associated with higher demand. This is quite intuitive as a major use case of bike sharing is to support first-mile/last-mile trips to/from mass transit systems. The structure of the road network around a BSS station also has a notable effect on its demand: BSS stations surrounded by more high-level road networks (e.g., motorway, primary) tend to have higher demand, while more low-level roads (e.g., residential) in the neighborhood can lead to lower demand. This might be because regions with high-level roads are usually associated with higher human flow and thus more potential customers. Meanwhile, the number of junctions in the neighborhood is negatively related to BSS demand, which might be due to the higher cost of riding in areas with more intersections and turns. POI density is another useful indicator of BSS demand. It is found that zones with higher residential, transportation and commercial POI density in the neighborhood can be associated with higher demand. This is reasonable as people can use bike sharing to access home or transportation facilities, and places with more commercial POIs are usually more prosperous and thus associated with higher demand. Socio-demographic features also affect BSS station demand significantly. Among them, the percentage of white residents in the census tract is one of the most important features, and areas with more white residents generally have higher BSS demand.

\subsection{Understanding spatial interactions between BSS stations}

\begin{figure}[ht!]
    \centering
    \includegraphics[width=\linewidth]{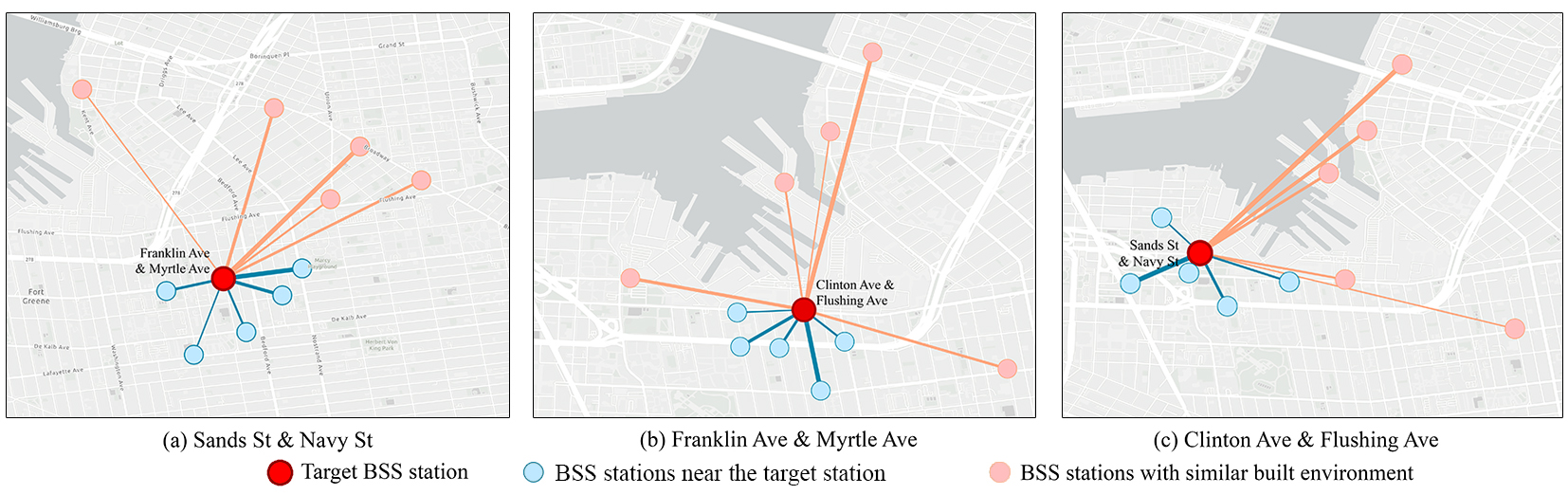}
    \caption{Example BSS stations and their spatial dependencies (with thicker dashed lines denoting higher attention weights)}
    \label{fig:GAT_visual}
\end{figure}

In addition to the relationship between the built environment and BSS demand, we can also use our model to understand the local spatial interactions between BSS stations by examining the attention weights learned from the GAT layers. Recall that the attention weights represent spatial dependencies of a target BSS station on other stations in its localized graph. In Fig.~\ref{fig:GAT_visual}, we choose three newly added stations in the test set as examples and show the learned attention weights between the target station and connected stations. Each dot represents a BSS station and a thicker dashed line between two dots represents a higher attention weight and thus stronger correlations between stations. It can be found that both the stations with close geographical proximity and those with high built environment similarity can have strong correlations with the target station. In addition, the interaction does not strictly follow the distance decay: stations that are further from the target station can contribute more to its prediction. This suggests the advantage of using attention mechanisms to learn station interactions instead of using pre-defined adjacency weights. 

\section{Conclusion} \label{sec:conclusion}

This research focuses on the trip generation problem for BSS expansion based on urban built environment characteristics. Previous research typically rely on regression or machine learning models, which might not be adequate to capture the nonlinear effect of the built environment or the complex spatial interactions between BSS stations. While deep learning methods have shown promise for such demand prediction tasks, most existing models focus on the short-term prediction for mature and stable systems, and are generally not applicable to trip generation for system expansion scenarios. To address these issues, this study introduces a graph deep learning approach for TG-BSSE. It leverages various built environment features as model inputs, including POIs, road network, transportation facilities, socio-demographics and BSS network design. To capture spatial interactions between BSS stations, we construct localized graphs centered on each target BSS station based on both geographical proximity and built environment similarity, and adaptively learn their spatial correlation weights using attention mechanisms. We further demonstrate that the proposed GNN approach can be seen as a generalized spatial regression model with nonlinear activation functions, heterogenous spatial dependencies and adaptive spatial weights, which allows us to synergize the development of GNN and spatial regression methods. Using a real-world BSS expansion dataset from NYC over multiple years, experiment results verify the improved performance of our proposed model compared to existing methods as well as alternative deep learning variants. Furthermore, we demonstrate the model interpretability regarding how different built environment features affect BSS demand and how BSS stations interact with each other. The proposed model can be used to support strategic planning for BSS expansion, especially related to the site selection and capacity design for new stations. The consideration of spatial dependencies across new and existing stations allow us to evaluate the potential impact of new stations on the whole system.

This research can be further improved or extended in several aspects. First, currently we only exploit urban built environment features to make predictions, and future research can explore how to better leverage both built environment and historical demand data for long-term transportation planning. For example, inspired by recent research for cold-start user recommendation \citep{dong2020mamo}, it is possible to use memory-augmented meta-learning approaches to transfer knowledge learned from historical demand of existing stations to newly planned ones. Second, although this research focuses on trip generation at the station level, it is also important to understand how the generated trips would be distributed across different OD pairs (i.e., the trip distribution step in four-step travel demand forecasting). One challenge is that the OD-level demand observations can be rather sparse, with most OD pairs having few trips, which leads to computational robustness issues. Nevertheless, recent development in uncertain quantification methods can be incorporated to mitigate such concerns \citep{zhuang2022uncertainty}. Third, currently our proposed model focuses on the use case for system expansion, in which the training and testing data are generated in the same city. This is no longer applicable when a city has to plan a BSS from scratch. One possible solution is to extend our approach for cross-city planning scenarios, where the model is trained in one city and deployed in another. The performance likely depends on many city-specific factors, some of which are not easily quantifiable (e.g., cycling culture, BSS branding). Last but not least, as BSS strategic planning is cited as one of the main target applications for our demand prediction model, a natural next step is to develop a larger framework to directly incorporate the prediction results with a joint optimization framework for station location selection and capacity design. This will serve as a valuable toolkit for the data-driven planning and design of future BSS as well as other transportation networks. 

\section*{Acknowledgements}
This research is supported by National Natural Science Foundation of China (NSFC 42201502) and Seed Fund for Basic Research for New Staff at The University of Hong Kong (URC104006019).

\appendix
\section{Comparison with DNNs based on sequential dependencies}\label{appendix:a}

As introduced in Section~\ref{literature:new}, existing deep learning approaches formulated TG-BSSE as a time-series prediction problem with historical demand patterns of existing stations as input. They mostly use simulation data for experiments with short time intervals (i.e., from 30 minutes to 1 day). To test their performance in real-world BSS expansion scenarios, we design a problem formulation as follows: assuming the planning and construction of BSS stations take $K$ months, to predict the potential demand for BSS expansion at time step $t$, we use the historical demand series of existing stations from month $t-K-T$ to month $t-K$ as model input. In the test data, we define stations that do not exist at time step $t-K$ as new observations and the others as existing observations. In our experiments, we set $T=6$ and $K=6$. This results in 2336 and 19472 new and existing observations for model evaluation respectively. We compare our proposed model with two existing baselines based on temporal sequential dependencies:

\begin{itemize} [noitemsep]
    \item \textbf{DDP-Exp} \citep{luo2019dynamic}: a graph sequence learning approach to trip generation for time-varying transportation networks, which uses LSTM to capture temporal dependencies and GCN to capture spatial dependencies. 
    \item \textbf{MOHER} \citep{zhou2021modeling}: a spatiotemporal framework to predict the number of trips generaged for a newly planned transportation site by aggregating the historical demand of multi-modal transportation sites that are either geographically adjacent or have similar POI distributions. We implement a simplified version considering only historical demand of BSS stations. 
\end{itemize}

\begin{table}[ht!]
  \centering \footnotesize
  \caption{Performance comparison with DNNs based on sequential dependencies}
    \begin{tabular}{ccccccc}
    \toprule
    \multirow{2}{*}{Models} & \multicolumn{3}{c}{New observations} & \multicolumn{3}{c}{Existing observations} \\
    & RMSE & MAE & $R^2$ & RMSE & MAE & $R^2$\\
    \midrule
    DDP-Exp & 39.913 & 35.804 & 0.325 & 43.472 & 30.920 & 0.621 \\
    MOHER & 33.125 & 24.733 & 0.663 & 41.831 & 26.180 & 0.653 \\
    Spatial-MGAT & \underline{\textit{22.375}} & \underline{\textit{13.305}} & \underline{\textit{0.783}} & \underline{\textit{27.495}} & \underline{\textit{18.095}} & \underline{\textit{0.848}}\\
    \bottomrule
    \end{tabular}
  \label{table:sequential_dnn}%
\end{table}%

The results displayed in Table~\ref{table:sequential_dnn} show that the aforementioned DNNs based on temporal sequential dependencies might not work well in real-world BSS expansion scenarios, likely for the following reasons. First, they assume that the demand data used for model training and test follows similar data distribution, neglecting the fact that the addition of new stations can lead to structural changes for the entire system. Second, with a longer and more realistic time interval (i.e., month) in our experiment setting, they may suffer from data sparsity issues and can easily lead to overfitting. Finally, in cases when newly planned stations are located far from existing ones, it is difficult to leverage historical demand from nearby sites for model fitting. This is often the case for real-world system expansion, in which a cluster of new stations in the same neighborhood are planned and deployed at the same time. Compared with these methods, we formulate TG-BSSE as a spatial regression problem by leveraging multi-source urban built environment features, which are more general and reliable for long-term transportation planning applications.






\bibliographystyle{model5-names2}\biboptions{authoryear}
\bibliography{ref}







\end{document}